\pdfoutput=1

\documentclass[11pt]{article}

\usepackage{EACL2023}

\usepackage{times}
\usepackage{latexsym}

\usepackage[T1]{fontenc}

\usepackage[utf8]{inputenc}

\usepackage{microtype}

\usepackage{multirow} 
\usepackage{multicol}
\usepackage{graphicx}
\usepackage{mathabx}

\usepackage{booktabs,arydshln}

\makeatletter
\def\adl@drawiv#1#2#3{%
        \hskip.5\tabcolsep
        \xleaders#3{#2.5\@tempdimb #1{1}#2.5\@tempdimb}%
                #2\z@ plus1fil minus1fil\relax
        \hskip.5\tabcolsep}
\newcommand{\cdashlinelr}[1]{%
  \noalign{\vskip\aboverulesep
           \global\let\@dashdrawstore\adl@draw
           \global\let\adl@draw\adl@drawiv}
  \cdashline{#1}
  \noalign{\global\let\adl@draw\@dashdrawstore
           \vskip\belowrulesep}}
\makeatother

\title{\textit{Made of Steel?}\\Learning Plausible Materials for Components in the Vehicle Repair 
Domain}

\author{Annerose Eichel, Helena Schlipf, Sabine Schulte im Walde \\
         Institute for Natural Language Processing, University of Stuttgart \\ 
         \texttt{\small \{annerose.eichel,schulte\}@ims.uni-stuttgart.de, helena.schlipf@gmail.com}}

\begin{document}

\maketitle

\begin{abstract}
We propose a novel approach to learn domain-specific plausible materials for components in the vehicle repair domain by probing Pretrained Language Models (PLMs) in a cloze task style setting to overcome the lack of annotated datasets. We devise a new method to aggregate salient predictions from a set of cloze query templates and show that domain-adaptation using either a small, high-quality or a customized Wikipedia corpus boosts performance. When exploring resource-lean alternatives, we find a distilled PLM clearly outperforming a classic pattern-based algorithm. Further, given that 98\% of our domain-specific components are multiword expressions, we successfully exploit the compositionality assumption as a way to address data sparsity.  
\end{abstract}

\section{Introduction}

\begin{figure}[t!]
\centering
  \includegraphics[width=0.93\linewidth]{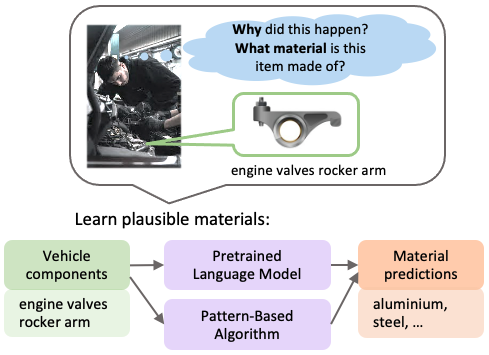}
  \vspace{-1mm}
  \caption{Conceptual overview for learning domain-specific plausible materials (e.g., \textit{aluminium}) for
  vehicle components (e.g., \textit{engine valves rocker arm}).}
  \label{fig:example}
\vspace{-6mm}
\end{figure}

Connecting a symptom to an underlying cause is a crucial building block for natural language understanding across domains. For example, as illustrated in Fig.~\ref{fig:example}, a standard approach that human mechanics in vehicle repair shops apply when tracing the cause of a symptom, is to exploit the link between a vehicle \textit{component} and the component's \textit{materials}. This study 
tackles the task of automatically learning plausible materials for vehicle components. The information is crucial in vehicle repair shops, as mechanics are faced with constantly growing vehicle complexity, making it hard to manually identify the cause of a malfunction. While domain-specific information on vehicle components is often available, plausible domain-specific material information typically needs to be gathered from external data sources. For example, 
from ``\textit{Brake disk are usually manufactured from gray cast iron}'', one may retrieve the information that \textit{brake disks} consist of 
\textit{iron} (and possibly further materials not mentioned here).

We propose a novel approach to leverage state-of-the-art pretrained language models (PLMs) for the task of learning plausible materials in highly domain-specific contexts, such as vehicle components in the vehicle repair domain. Our approach can also be applied to similarly specific domains, e.g., to learn which materials or substances are plausibly used in products in the health domain, textile industry, etc. 
As we tackle a task from material science, we make a contribution to a field that is not only very challenging, but heavily under-investigated from an NLP perspective.
We focus on the PLM RoBERTa \cite{liu2019} and develop an approach that successfully learns plausible materials in English for components in the vehicle repair domain. 
We further exploit domain-adaptation variants and show (i)~that harnessing a small high-quality domain-specific corpus boosts the performance over an out-of-the-box vanilla RoBERTa, and (ii)~how similar performance can be reached by using a widely accessible data source such as Wikipedia if no domain-specific data is available.

To overcome the typical lack of annotated datasets for training and fine-tuning models 
in highly domain-specific tasks, we probe RoBERTa in a cloze task setting \cite{taylor1953,petroni-etal-2019-language},
compare variants for aggregating the results, and analyze the influence of a varied set of query templates on model predictions. 

While PLMs come with great advantages, their hunger for storage, data, and computing power is even greater. We thus compare the performance of the vanilla and domain-adapted versions of RoBERTA with (i)~a distilled version, i.e., the much smaller DistilRoBERTA \cite{sanh2019}, and (ii)~the seminal pattern-based bootstrapping algorithm Basilisk \cite{thelen-riloff-2002-bootstrapping}.

Finally, we tackle the challenge of handling a dataset with a vast majority of multiword expressions, i.e., 98\% of our targets are noun compounds (such as  \textit{seat heating switch led}).
We address the corresponding severe data sparsity by assuming that many domain-specific compounds are compositional \cite{clouet2014splitting,HaettyEtAl:21} such that our domain-specific model may fall back to information regarding the compound's head (in the example above: the right-most simplex noun in the compound \textit{led}), and thus improve on the data sparsity.

\section{Related Work}

\paragraph{Domain-specific IR}
In contrast to domains such as biomedicine and chemistry, ``[l]everaging NLP tools in materials science remains in its infancy'' \citep[p.4]{olivetti2020}, with main challenges including the development of task- and domain-specific tools
to structure and harnes knowledge for material synthesis and material discovery. Common approaches to elicit material mentions from text include unsupervised approaches such as topic modeling \citep{rani2021} and LDA \cite{glass2021}, as well as supervised methods for domain-specific NER, often focusing on a single material or material group \citep{mysore2017,mysore-etal-2019-materials, friedrich-etal-2020-sofc,gupta2022,anmol2021,ogorman-etal-2021-ms}. Differently to previous work,
we probe PLMs in a cloze-task style setting and compare the results to a similarly unsupervised pattern-based bootstrapping algorithm. While most unsupervised methods are not easily interpretable, our query templates, seed words, patterns and predicted materials are directly accessible.


\paragraph{Prompting and Cloze Query Engineering}
Recent work has leveraged cloze-task style settings to probe the knowledge that PLMs acquire during pretraining, thus targeting linguistic capabilities \cite{goldberg2019,ettinger-2020-bert,apidianaki-gari-soler-2021-dolphins,rogers2020}, the understanding of rare words \cite{schick2020} and conceptual abstractions \cite{ravichander-etal-2020-systematicity}, as well as factual and commonsense knowledge \cite{petroni-etal-2019-language,jiang-etal-2021-know}. An emerging strand of research highlights PLM sensitivity to the input in a cloze-task style setting. For example, \citet{elazar-etal-2021-measuring} demonstrate that PLMs show deficiencies in prediction consistency when presented cloze-style query paraphrases. Others exploit this flaw by explicitly considering paraphrases of a cloze query in addressing a model's sensitivity to a specific input query to elicit a desired output \cite{davison-etal-2019-commonsense,jiang-etal-2021-know}.
\citet{pandia-ettinger-2021-sorting} show that models lack robustness in their ability to harness relevant context information in the face of cloze tasks containing distracting contextual cues. In reverse, priming the model using \textit{trigger tokens} \cite{shin-etal-2020-autoprompt} and \textit{lexical cues} \cite{misra-etal-2020-exploring} might aid in guiding the model to predict the desired output. Specifically, \citet{shin-etal-2020-autoprompt} propose \textsc{AutoPrompt} to develop automatically constructed prompts or patterns to elicit knowledge from pretrained PLMs for a variety of tasks.
In contrast to previous work, we semi-automatically construct a variety of prompts to cover a salient set of paraphrases for eliciting plausible materials in a PLM; this allows for meaningful comparison and interpretation of the predictions' quality and the influence of the cloze queries.

\paragraph{Semantic Plausibility} While classical distributional models tend to model selectional preferences and thematic fit instead of capturing semantic plausibility \cite{erk-etal-2010-flexible}, we have recently seen advances to model plausibility across various dimensions, including \textit{physical} and \textit{abstract} semantic plausibility \citep{wang-etal-2018-modeling,porada-etal-2019-gorilla}. SOTA models for event plausibility however still rely on straightforward conditional probabilities of co-occurrences as estimated by distributional models \citep{emami-etal-2021-adept,porada-etal-2021-modeling}.
Considering our task of learning plausible materials for vehicle components, we go beyond selectionally preferred component materials as predicted with high probabilities, aiming also for less frequently observed cases that are nevertheless plausible.

\paragraph{(Domain-specific) MWEs and Compositionality} Multiword expressions (MWEs) are challenging for any natural understanding system, given that MWE meanings are idiosyncratic to some degree, i.e., the meaning of an MWE is not entirely (or even not at all) predictable from the meanings of the constituents \cite{SagEtAl:02,ReddyEtAl:11a,SalehiEtAl:14b,SchulteImWaldeEtAl:16b,CordeiroEtAl:19,SchulteImWalde/Smolka:20}. Even though MWEs are ubiquitous not only in general- but also in domain-specific language \cite{clouet2014splitting,HaettyEtAl:21}, up to date only few NLP systems have exploited MWE meaning modules, as in machine translation \citep{Cholakov_Kordoni:14,WellerEtAl:14b}.
This study is faced with 98\% noun compounds among our domain-specific targets, and we test the compound--head compositionality assumption (e.g., a \textit{seat heating switch led} "is a type of" \textit{led} but an \textit{engine valves rocker arm} "is \underline{not} a type of" \textit{arm}) 
to fight the severe MWE-triggered data sparsity.

\section{Data} \label{sec:data}


\paragraph{Vehicle Component Dataset} As targets for our components, we rely on a set of 7,069 unique component names curated by experts from the vehicle repair domain.\footnote{The Vehicle Component Dataset and the DOMAIN corpus are provided by Bosch.} A component name may denote a tangible physical component such as  \textit{cooling blower}, as well as intangible functional and software components such as \textit{ABS warning lamp function} and \textit{road test}.
The dataset comprises 155 single-word components and 6,914 multiword components with up to eight constituents (see Fig.~\ref{fig:vcd_mwe_distribution} in App.~\ref{sec:data_distribution} for the distribution). When applying the assumption of MWE compositionality, we fall back to the right-most constituent word as the MWE head
\cite{altakhaineh2019}, resulting in a total of 725 different heads across the 6,914 multiword components.

\paragraph{Domain-Specific Corpora}

We utilize two domain-specific English corpora for extracting materials for vehicle components from text: (i) a domain-specific vehicle repair manual written by domain-experts, containing approx. 
800K English tokens (henceforth: DOMAIN) and (ii) a portion of the English Wikidata (henceforth: WIKI). For this,
we download a Wikidata dump\footnote{\scriptsize{\url{https://dumps.wikimedia.org/enwiki/}}, 2.3.2022}, from which we only keep those 118,154 articles which contain one or more of the 6,914 multiword components in the titles or body
\cite{hatty-etal-2020-predicting}.
For domain-adaptation, we delete sentences that do not contain any of the multiword components, resulting in approx. 225K sentences.

\section{PLM Experiments}

\begin{figure*}[htbp!]
\centering
  \includegraphics[width=0.99\linewidth]{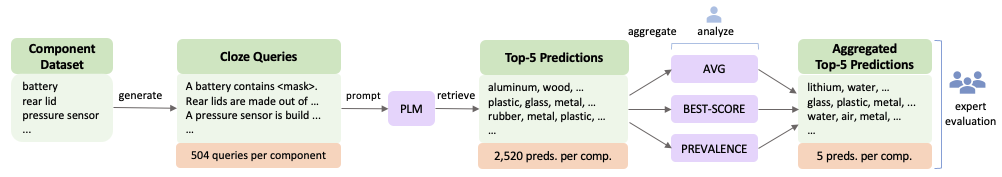}
  \caption{Overview of prediction generation and aggregation, with the numbers in the orange boxes referring to queries/predictions for \textit{one} example component, e.g., \textit{battery}.}
  \label{fig:aggregation_overview}
\vspace{2mm}
\end{figure*}

\begin{table*}[!h]
\centering
\small
\begin{tabular}{ll}
\toprule
\multicolumn{2}{l}{\textbf{Cloze Query Templates}}                                                                                          \\
\multicolumn{2}{l}{{[}context]~+~[noun\_quant.]~+~indef. article~+~component~+~verb\_relation + [adverb]~+ \textless{}mask\textgreater{}.}    \\ 
\midrule
\textsc{singular}~~                      & a battery consists of <mask>.~                                                                                      \\
\textsc{plural}~~~~                      & batteries consist of <mask>.~~                                                                                     \\
noun\_quantifier + \textsc{plural}~~~~   & most batteries consists of <mask>.~~                                                                                \\
adverb + \textsc{singular}~~~~ & a battery usually consists of <mask>.~~                                                                             \\
adverb + \textsc{plural}~~~~   & batteries usually consists of <mask>.~~                                                                             \\
context + \textsc{singular}~~~~~         & when used in a vehicle, a battery consists of <mask>.~                                                              \\
context + \textsc{plural}~~~             & when used in a vehicle, batteries consists of <mask>.~                                                              \\
\bottomrule
\end{tabular}
\caption{Cloze query template and filled-in queries for the vehicle component \textit{battery}. \textsc{singular} and \textsc{plural} are placeholders that are filled in with vehicle component nouns in singular and plural form, respectively.}
\label{tab:cloze_statements}
\vspace{-3mm}
\end{table*}

Our PLM experiments investigate 
to what extent and how PLMs encode domain-specific component materials when prompted with cloze query variants, as illustrated in Fig.~\ref{fig:aggregation_overview}: We use the vehicle component dataset to construct a set of 504 cloze query templates for each individual component. We first probe a RoBERTa model without any modifications and gather the top-5 predictions for each component, resulting in a total of >2K predictions. To aggregate these predictions, we experiment with three aggregation methods and determine the top-5 most plausible predictions per component and aggregation method for expert evaluation.
In a follow-up experiment, we perform domain-adaptation of RoBERTa and DistilRoBERTa for the vehicle repair domain using our two domain-specific corpora, and 
compare the probing results of the vanilla and the domain-adapted models. The following subsections describe our PLM experiments in detail.

\subsection{Cloze Query Prediction and Processing} \label{subsec:cloze_query_prediction_processing}

Our cloze query handling involves three steps (again, cf. Fig.~\ref{fig:aggregation_overview}): we first construct a set of cloze query templates, then we aggregate model predictions, and finally we analyze the effect of individual cloze queries to select the most meaningful templates for the final top-5 predictions. As Vanilla RoBERTa model \cite{liu2019} (henceforth: Vanilla RB), we draw on the \texttt{roberta-base} implementation by \texttt{huggingface} \cite{wolf-etal-2020-transformers} without any modifications.


\paragraph{Step 1: Cloze Query Template Construction} \label{para:cloze_query_construction}

We develop a set of templates and generate cloze statements for each vehicle component to probe our PLM for plausible component materials:
First of all, we define a set of 18 paraphrases typically used to express that a component is made from one or more materials \cite{davison-etal-2019-commonsense,jiang-etal-2021-know}. For this, we start with a set of highly frequent verb relations in our corpus (such as \textit{contain}) and use WordNet synsets \cite{Fellbaum:98} to determine additional relevant verbs.
We then apply both plural and singular forms of the component nouns with corresponding indefinite articles to define queries
(see Table~\ref{tab:cloze_statements}).
To include prompts that refer to prototypical (and thus presumably highly plausible) materials, we follow \citet{apidianaki-gari-soler-2021-dolphins} and integrate the quantifiers \{\textsc{most}\} and \{\textsc{all}\}; we also add the quantifier \{\textsc{many}\}.
Furthermore, we define a query template element for adverbs referring to typicality, by including the adverbs \{\textsc{usually}, \textsc{generally}, \textsc{normally}\}. Fostering materials that are not necessarily prototypical but still potentially plausible, we use the noun quantifier \{\textsc{some}\} \cite{apidianaki-gari-soler-2021-dolphins} and the adverbs \{\textsc{possibly}, \textsc{plausibly}\}.   
In our domain-specific scenario, 
humans tend to intuitively restrict the answer space of plausible materials by providing contexts such as ``\textsc{When used in a vehicle, ...}''. 
Accordingly, we follow recent work investigating the impact of so-called \textit{trigger tokens} \cite{shin-etal-2020-autoprompt} and \textit{lexical cues} \cite{misra-etal-2020-exploring}
to predict a related token, and define six limiting context phrases to leverage intuitive human behaviour.

The above procedure constructs 504 cloze queries for each vehicle component such as the input query ``\textit{a battery contains <mask>.}'', where the plausible material candidate is masked and to be predicted by a PLM. 
A full overview of query variants is presented in Table~\ref{tab:cloze_statements}.

\paragraph{Step 2: Cloze Query Prediction Aggregation} \label{para:aggregating_query_predictions}

For each component, we prompt our PLM with the corresponding 504 cloze queries and obtain the top-5 predicted tokens ranked by probability for each of the queries, resulting in 2,520 predictions per component, see Fig.~\ref{fig:aggregation_overview}.\footnote{We apply basic post-processing as described in App.~\ref{subsec:postprocessing}.}
Given that we want only a small list of highly plausible materials, we experiment with three approaches to aggregate the 2,520 predictions for each component such that the most plausible material candidates are ranked at the top of a component's material list.

\begin{itemize}
    \item \textsc{best-score} probabilities aggregate the most probable PLM-predicted material types from the 2,520 top-5 predictions across all queries. 
    In this way, predictions that the model considers highly probable in a specific query constellation 
    are considered highly plausible, no matter how often they have been predicted.
    \item \textsc{avg} probabilities are obtained by summing  up the probabilities for each query variant
    on the material type level and then averaging by the number of queries that predicted that
    material.
    This way, probability mass is taken away from 
    materials that were predicted with high probabilities by only one or few queries. The aggregation is thus directed towards materials that are less prominent for individual queries but more pervasive across query variants.
    \item \textsc{prevalence} refers to a method where the most salient PLM-predicted materials are obtained by ranking only according to how often the material was predicted across all top-5 suggestions of all query variants. Completely ignoring the probabilities in this step accounts for the amount of used cloze queries and provides insight into whether the model leverages semantic information across specific queries.
\end{itemize}

\begin{table}[!htbp]
\begin{small}
\centering
\begin{tabular}{lrrr}
\toprule
                     & \textsc{avg} & \textsc{best-score} & \textsc{prevalence}  \\
\midrule
Total@5    & 0.08                       & 0.18                              & \textbf{0.42} \\
Acc@5  & 0.36        & 0.58                             & \textbf{0.86}        \\                    
\bottomrule            
\end{tabular}
\caption{Results of aggregation methods on top-5 predicted materials for 100 components each, i.e., a total of 500 predictions per method, reporting total@5 and accuracy in \%.}
\label{table:aggregations}
\vspace{-3mm}
\end{small}
\end{table}

After aggregating the final top-5 predictions by using each of the three presented methods, we analyze\footnote{The analysis is performed by one of the authors of this paper, who is familiar with the vehicle repair domain. Note that this preliminary assessment is only for comparing aggregation methods, and therefore independent of the final evaluation involving three experts, see Fig. 2.}
for 100 components (see App.~\ref{sec:component_sampling} for sampling details) whether the top-5 material predictions are plausible.
We follow previous work \cite{ettinger-2020-bert,apidianaki-gari-soler-2021-dolphins} and evaluate by accuracy, i.e. the proportion of vehicle components for which a suggested material is among the model's top-5 predictions.

Results are reported in Table~\ref{table:aggregations}. The suggested materials ranked by \textsc{prevalence} with over 40\% of correct predictions
clearly outperform \textsc{best-score} and \textsc{avg}.
Especially \textsc{avg} is observed to promote rather implausible predictions such as \textit{dna}, \textit{food}, or \textit{death}.
Based on this preliminary assessment, we use the \textsc{prevalence} method for aggregating the top-5 predictions in our final experiments.

\paragraph{Step 3: Cloze Query Template Selection} \label{para:analysis_selection}

Using the top-5 material predictions by \textsc{prevalence}, we now investigate the productiveness of query templates for 100 components (see App.~\ref{sec:component_sampling} for sampling details) in order to select the most salient ones for our final experiments. Productiveness investigates which queries are \textit{productive}, i.e., contribute a material to the final top-5 materials.

\begin{figure}[htbp!]
\centering
  \includegraphics[width=\linewidth]{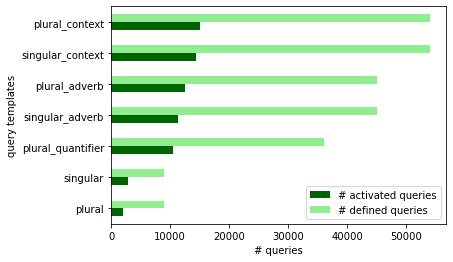}
  \vspace{-5mm}
  \caption{Query template productiveness
  across the respective defined query template variants.}
  \label{fig:freq_qv}
\vspace{-0mm}
\end{figure}

In our first productiveness analysis, we analyze the 68,651 
queries that trigger any of the top-5 
predictions across all 100 sampled components and all query template variants. 
Fig.~\ref{fig:freq_qv} shows the number of queries actively triggering a top-5 prediction in comparison to the overall number of queries that could have potentially been activated (in absolute numbers).
Overall, plural variants exhibit a slightly higher productiveness than singular variants. 

\begin{figure}[htpb!]
\centering
  \includegraphics[width=\linewidth]{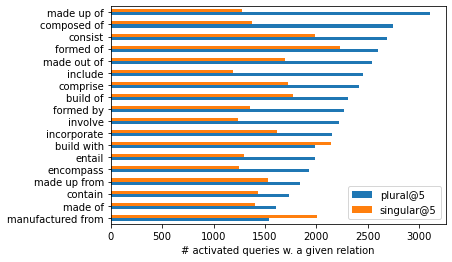}
  \vspace{-5mm}
  \caption{Relation productiveness across singular and plural query template variants.}
  \label{fig:freq_rel5}
\vspace{-3mm}
\end{figure}

In our second productiveness analysis, we examine whether some verb relations "do all the work" in comparison to others that are not activated and should therefore be excluded for the final experiments.
Fig.~\ref{fig:freq_rel5} shows the relations triggering a top-5 prediction across all query variants. The most productive relations for singular and plural variants
are [\textit{comprise, formed of, build with}] and [\textit{made up of, composed of, consist of}], respectively.
For all but two cases, query productiveness of plural variants is higher than for the singular equivalents.

Our two productiveness analyses suggest that our set of domain-specific query templates represents a well-defined set 
which is \textit{successful as a whole} in cooperation with a suitable method for aggregating a large number of resulting predictions, and does not include strongly over- or underperforming variants. For our final PLM experiments, we thus leverage the full set of query templates.

\subsection{Domain Adaptation of RoBERTa} \label{subsec:domain_adaptation}
We experiment with our two domain-specific corpora DOMAIN and WIKI (see §\ref{sec:data}) to adapt RoBERTa to the vehicle repair domain (henceforth: DOMAIN RB and WIKI RB, respectively).
Additionally, we compare RoBERTa to a domain-specific resource-lean model DOMAIN DistilRB, where we adapt DistilRoBERTa \cite{sanh2019} to the vehicle repair domain using DOMAIN.

The domain adaptations draw on the \texttt{roberta-base} and \texttt{distilroberta-base} implementations from \texttt{huggingface} \cite{wolf-etal-2020-transformers}; we split our input texts into train (90\%) and validation (10\%) sets. 
Following common choices,
we set the 
argument specifying 
the fraction of tokens to mask to 15\%. We train with a learning rate of 2e-5, a weight decay of 0.01, and a batch size of 1,024 (cf. App.~\ref{subsec:domain_adaptation} for further details). We train for three epochs and save the best model according to validation set performance. We train one model for each of the aforementioned variants (DOMAIN RB, WIKI RB, and DOMAIN DistilRB). Model results will be presented in §\ref{sec:results}, in comparison to Basilisk results.


\section{Basilisk Experiments:
A Classic}

In comparison to the PLM experiments, we explore the seminal pattern-based bootstrapped learner Basilisk \cite{thelen-riloff-2002-bootstrapping} for learning domain-specific plausible materials for vehicle components. Basilisk is designed to learn high-quality semantic lexicons for one or more categories, provided an unstructured natural language text corpus plus seed words for each semantic category to be learnt. The resource-lean learner leverages extraction patterns representing linguistic contexts, thus exploiting lexico-syntactic structures to capture word meaning. 

As the starting step for Basilisk, we select ten seed nouns from the most frequent words in
DOMAIN\footnote{Set of seeds: \{\textit{water, steel, metal, glass, rubber, plastic, aluminum, copper, polyester, quartz}\}
In the beginning, we created individual seed word lists for DOMAIN and WIKI, where the latter lacked sufficient quality. Thus, we harness the DOMAIN seed words for both datasets.
} that belong to the 
semantic class \textit{material}. 
Since Basilisk depends on extraction patterns to supply contextual support for additional words belonging to the same semantic class, we rely on dependency parsing to create domain-specific syntactic contextual patterns by using the seed words obtained from DOMAIN and only the DOMAIN corpus.
We then apply Basilisk as bootstrapping algorithm to select the best-performing patterns for a pattern pool and subsequently fill a candidate word pool with the extractions of the highest-scoring patterns. We retrieve, for example, 
\texttt{\textsc{seed}*\footnotesize{<GDep>:<compound>:<dependent>:alloy}}\\ 
on the basis of which the word \texttt{alloy} gets added to the candidate word pool. We stop Basilisk after \textit{n} bootstrapping rounds and store the retrieved semantic lexicon of \textit{n} plausible materials for a given vehicle component. 
To then connect these possible materials with the components, we first process the underlying unstructured data sources and only keep sentences containing at least one component name mention. Second, we filter the obtained sentences with the semantic lexicon and store all component-material candidate matches.


In this way, we generate 396,887 extraction patterns,
and use the Basilisk implementation 
\citep{thelen-riloff-2002-bootstrapping} with all corresponding patterns and seed words on the DOMAIN dataset. We limit the candidate word pool size to $n=200$,
and generate possible material candidates for both the DOMAIN and the WIKI datasets. 
\section{Evaluation} \label{sec:evaluation}
To our knowledge, no gold standard is available for the task of plausible material extraction in the vehicle repair domain. We thus perform an expert evaluation of the quality of predictions for each slot obtained from the vanilla and the domain-adapted RoBERTA models. To compare PLM-based results to resource-leaner methods, we also evaluate results from the pattern-based Basilisk algorithm and 
DistilRoBERTa.

\paragraph{Evaluation Task}
For each model, we select and evaluate 100 randomly sampled components,\footnote{We make the list of evaluated components available
\href{https://github.com/AnneroseEichel/EACL2023_made-of-steel}{here}.} (i)~balancing the number of constituents per target component,
(ii)~providing the full component vs. providing only its head, and (iii)~varying the underlying corpus (for Basilisk only). For both 1- and 2-constituent words,\footnote{A 1-constituent word is a simplex word.} we sample 30 components with 20 full/10 head samples with equal amounts from DOMAIN/WIKI. For 3-constituent words, we sample 20 instances with equal amounts of full/head and DOMAIN/WIKI samples. For both 4- and 5-constituent words, we sample 10 head instances with equal amounts of DOMAIN/WIKI samples (cf. App.~\ref{sec:component_sampling} for further details).

We obtain the aggregated top-5 predictions for vehicle components from each model. 
For Basilisk, we sample up to five answer options whenever more than five entries are extracted for a given component. 
\begin{table}[!htpb]
\centering
\small
\begin{tabular}{lr|rrr}
\toprule
Model            & 
IAA & A1::A2 & A2::A3 & A3::A1\\ \midrule
Vanilla RB      & 0.85    & 0.89     & 0.81     & 0.86      \\ 
DOM. RB       & 0.79    & 0.85     & 0.74     & 0.78      \\
WIKI RB         & 0.81    & 0.86     & 0.74     & 0.83      \\
Basilisk        & 0.79    & 0.79     & 0.74     & 0.83      \\
DOM. DistilRB & 0.82    & 0.88     & 0.78     & 0.81      \\ \bottomrule
\end{tabular}
\vspace{-1.5mm}
\caption{Averaged (left) and pair-wise (right) IAA on material predictions.}
\label{table:IAA}
\vspace{-3mm}
\end{table}
To minimize annotator bias, we compile a set of up to 18 materials for each vehicle component while keeping track of prediction origins. This way, annotators see each evaluated component only once and rate all predictions from all models for each vehicle component without potential unconscious comparison to previously seen predictions from another model.
In the majority of cases, the models suggested materials either as a singular or as a plural form (e.g., \textit{metal} and \textit{metals}); for the few cases were we are faced with both versions, we decide to not merge the model predictions and ask the annotators to tick both options if both are considered plausible.

\paragraph{Evaluation Setup}

We present each component instance with corresponding materials to three experts\footnote{The participating annotators are not affiliated with the company providing the Vehicle Component Dataset.}
from the vehicle repair domain. 
The annotators are asked to rate the plausibility of material predictions for each vehicle component, by ticking the correctly identified materials. They are also provided the options ``none of these'' if no material is plausible, and ``I do not know the answer'' if this is the case. Inter-annotator agreement (IAA) is shown in Table~\ref{table:IAA}, with overall average IAA on the left and IAA calculated for each annotator pair on the right. All IAA scores indicate substantial agreement.
Further details on the annotation setup and on inter-annotator analyses on both material and component level are provided in App.~\ref{app:expert_eval}.

\section{Results and Discussion} \label{sec:results}


Table~\ref{table:expert_accuracy} presents results on model performances. On the left, we show accuracy on component level across $n$ annotators. Accuracy is defined as the proportion of components for which at least one of a model's top-5 material predictions is rated plausible by the annotators. On the right, we display accuracy on material level where accuracy is defined as the proportion of plausible material predictions among a model's top-5 predictions, as rated by $n$ expert annotators.\footnote{Material accuracy is naturally lower than component accuracy, as we evaluate five material predictions per component. Hence, achieving perfect agreement for material predictions is more difficult than perfect agreement for components.}

\begin{table}[]
\centering
\begin{small}
\begin{tabular}{lrr|rr}
\toprule
                 & \multicolumn{2}{r}{\textsc{components}}          & \multicolumn{2}{r}{\textsc{materials}} \\ \midrule
       Model          & $\geq$1A                  & 3A & $\geq$1A                  & 3A \\ \midrule
Vanilla  RB      & 0.87                           & 0.68   & 0.49                           & 0.24   \\
DOMAIN  RB       & \textbf{0.93} & \textbf{0.73}   & \textbf{0.62} & \textbf{0.28}   \\
WIKI  RB         & 0.91                           & 0.66   & 0.56                           & 0.24   \\  \cdashlinelr{1-5}
Basilisk         & 0.73                           & 0.40   & 0.45                           & 0.14   \\
DOMAIN DistilRB & 0.87                           & 0.69   & 0.53                           & 0.23   \\ \bottomrule
\end{tabular}
\caption{Model performance. \textsc{Component} accuracy: proportion of components for which at least one material among a model's top-5 predictions is rated plausible by $n$ expert annotators $n$A. \textsc{Material} accuracy: proportion of material predictions among a model's top-5 predictions rated plausible by $n$ expert annotators $n$A.
}
\label{table:expert_accuracy}
\vspace{-3mm}
\end{small}
\end{table}

\paragraph{Domain-Adaption Approaches: Vanilla, DOMAIN, and WIKI RoBERTa}
RoBERTa adapted to the domain using a small high-quality corpus (DOMAIN RB)
beats all other models with an accuracy of 0.93,
and RoBERTa adapted to the domain using a customized portion of the English Wikipedia (WIKI RB) is similarly successful (0.91); in comparison, Vanilla RoBERTa reaches an accuracy of 0.87. With perfect agreement among annotators (i.e., 3A), DOMAIN RB is still able to reach an accuracy of 0.73, in comparison to 0.66 for WIKI RB and 0.68 for Vanilla RB.
The top results on material level resemble those for component accuracy: DOMAIN RB (0.62) outperforms all other models, followed by WIKI RB (0.56). In contrast to the component level, Vanilla RB is no longer on par with the smaller but domain-adapted DOMAIN DistilRB, while still outperforming Basilisk. 
The differences between the three resource-intense approaches (Vanilla RB, DOMAIN RB and WIKI RB) are not significant\footnote{Significance tests apply $\chi^2, p<0.05$.} on component level for $n$A. On material level, however, differences are significant between PLMs (incl. DOMAIN DistilRB) for $\geq$1A, but not for 3A. All approaches significantly outperform the resource-lean Basilisk on both 
levels for all $n$A.

As a particular example to illustrate the positive impact of domain adaptation, we inspect results for the vehicle component \textit{engine valves rocker arm} from our introductory example (cf. Fig~\ref{fig:example}). 
Vanilla RB results include [\textit{wood, metal, steel, bones, legs}], where the predictions \textit{bones} and \textit{legs} refer to (parts of) an extremity of an animate being instead of a vehicle component. The materials [\textit{wood, metal, steel, joints, aluminium}] as obtained from WIKI RB instead encompass the material \textit{aluminium} as well as \textit{joints} which refer to connections between body parts as well as vehicle parts. Finally, DOMAIN RB results comprise all plausible materials predicted by the other models, namely [\textit{steel, metal, parts, aluminium, plastic}]. Moreover, the results feature the additional in-domain material \textit{plastic}, while not including out-of-domain predictions such as \textit{wood} anymore. 
We note that annotators are quite likely to rate one of the top-5 predictions for the same component as plausible, however, material prediction accuracy values indicate that the likeliness to also agree on this specific material drops with an increasing number of annotators.

The findings from these results are twofold. Domain adaptation of a given model results in a clear increase of plausible materials for items within a given domain. Second, if no highly domain-specific and well-curated corpus is available, 
domain adaptation using WIKI leads to very viable results and therefore represents a strong alternative.

\paragraph{Resource-Lean Approaches}
Results for the pattern-based 
algorithm Basilisk significantly underperform all other models for both component (0.73) and material (0.45) accuracy. In contrast, the domain-adapted resource-lean DistilRoBERTa (DOMAIN DistilRB) is on par with Vanilla RB and WIKI RB.
Manually analyzing predictions for our example discussed earlier, i.e., the component \textit{engine valves rocker arm}, we find that the Basilisk results [\textit{structures, glass, core, steel, plating}] are potentially within the domain, but often neither a material (\textit{structures, core}) nor rated plausible for the respective target component (\textit{glass, plating}). 
Results for DOMAIN DistilRB, in contrast, include [\textit{steel, metal, parts, plastic, aluminium}] and basically equal DOMAIN RB predictions.

The comparison of larger vs. leaner and smaller approaches by the examples of the well-established classic Basilisk and the transformer-based DistilRoBERTa model thus clearly demonstrates a preference for the latter.

\paragraph{Compositionality of Domain-specific MWEs}

\begin{figure}[!htbp]
\centering
  \includegraphics[width=\linewidth]{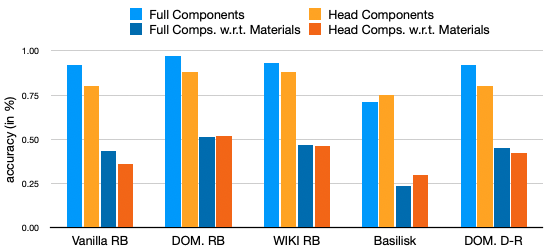}
  \caption{Comparison of full and head component performance on component (full: x/59, head: x/40) and material (full: x/395, head: x/200) accuracy level.}
  \label{fig:compositionality_analysis}
\vspace{-3mm}
\end{figure}

Results regarding the assumption of compositionality on domain-specific MWEs are presented in Fig.~\ref{fig:compositionality_analysis}. 
While accuracy values are lower when using the right-most head instead of the full multiword component (exception: Basilisk), the average difference is only 0.07 and 0.01 for component and material prediction accuracy, respectively.
Note, however, that IAA differs especially on material level with an average agreement of 0.44 using full and 0.32 drawing on head components. This indicates that predictions for full component names tend be more clearly connected to this very item than predictions retrieved for a head.

We further analyse model performance regarding the influence of MWE constituent number.
The results indicate that specificity beats brevity when learning plausible materials for vehicle components. However, our analysis also point towards a threshold separating a beneficial number of constituents (2) from a number pointing towards a detrimental degree of specificity (3+, cf. App.~\ref{subsec:n_tokens} for further details).

Our findings suggest that domain-specific multiword components are largely compositional and therefore justify a simplified meaning representation via head constituents.

\paragraph{The Influence of Cloze Queries}
To shed more light on the impact of using cloze query templates for material prediction, we examine whether the number of queries triggering predictions\footnote{An average number of 718 queries is activated for each component across models, with numbers of queries ranging from a minimum of $<$250 and a maximum of $>$1.200 queries activated per component.}
for a given component correlates with component accuracy. We find moderate-to-strong correlation values for all our PLMs, thus suggesting that a high number of queries yielding the same prediction indicates the prediction's plausibility.
We also test whether this correlation is directed and fit a corresponding linear regression model.
However, the resulting significant $R^{2}$ values, ranging around 0.25, are relatively weak and leave us with a mixed picture regarding a directed relationship between the number of queries and component accuracy (cf. App.~\ref{subsec:statistical_analysis} for further details).

When zooming into each template part, we find the adverb \textsc{generally} outperforming the other adverbs. Concerning quantifiers, we observe \textsc{many} yielding the most activated queries across models, followed by \textsc{most} and \textsc{all}. Among contexts, we identify \textsc{When used in vehicles,...} demonstrating best performance.
When inspecting the impact of our domain-specific verb relations, we find mostly distinctive groups of verbs accumulating at the extremes of the curve, e.g., \textit{build with, consist of, made up of} and \textit{involve, made of, include} yielding maximum and minimum numbers of plausible predictions across models and singular/plural variants, respectively. While overall observations suggest that templates in the passive voice produce more plausible predictions overall, examples such as \textit{consist of} underline the importance of a well-rounded template set that is \textit{successful as a whole}. 


To further explore the impact of designing high-quality query templates for automatic query construction, we perform the following experiment. We take the five most frequent main verbs \textit{make, say, go, use, take} from ENCOW16AX \cite{schaefer2015} that are considered to be \textit{not} directly related to describing what material an item is made of. We construct a set of cloze queries to probe our best-performing domain-adapted PLM DOMAIN RB and aggregate predictions as described in §\ref{subsec:cloze_query_prediction_processing}. We analyze the final top-5 material predictions and find only 9 correct materials. Instead, we observe a small number of types being predicted for many of the sampled components, 
as well as a small number of component-specific predictions that might be semantically related, but are no material, e.g., top-5 predictions for the component \textit{fuel tank} include [\textit{sense, noise, hydrogen, oil, cold}]. 
These findings emphasize that using a set that is (i) comprehensive, (ii) domain-specific, and (iii) syntactically diverse is crucial in order to avoid distortion of results.

\paragraph{Exploring Prediction Pool Size} Finally, we explore whether aggregating results from a larger pool of predictions obtained from the model benefits our goal of learning all plausible materials for a component. For this, we use our original set of 504 cloze queries per component to obtain the top-10 instead of previously top-5 predicted tokens from the PLM Vanilla RB, resulting in a set of 5040K predictions per component. We aggregate this doubled amount of predictions as explained in §\ref{para:aggregating_query_predictions} and analyze the final top-5 predictions as outlined in §\ref{para:analysis_selection}. While component accuracy differs by only 1\% (in favor of a smaller pool), drawing from a larger pool seems to benefit learning plausible materials with an 5\% increase of correct predictions when using top-10 model predictions for aggregation (cf. ~App.~\ref{subsec:pool_size_analysis} for details). We attribute these gains to the increase in relevant material predictions that are appropriately aggregated using \textsc{prevalence}, thus displacing, for example, non-relevant material predictions such as \textit{wood} from top-k positions.\footnote{Statistics w.r.t. \textit{wood} as a top-1 material prediction with pool size 5: 0.36\% (correct: 0.1\%); with pool size 10: 0.23\% (correct: 0.1\%).}

\section{Conclusion}

We tackled the task of learning domain-specific plausible materials for components in the vehicle repair domain from a novel perspective by probing SOTA language models in a cloze task style setting to overcome the lack of annotated datasets. Based on a diverse set of semi-automatically constructed cloze queries combined with a suitable aggregation method, we presented a new method to efficiently extract knowledge regarding vehicle components and their materials as acquired by a PLM. 
While showing that domain-adaptation using either a small, high-quality or a customized Wikipedia corpus boosts performance,  we also demonstrated the power of resource-lean alternatives such as the PLM DistilRoBERTa, and found that the bottleneck for domain-adaptation with respect to our task might not be model size but rather corpus quality and suitability. 
Finally, we successfully exploited the compositionality assumption for our domain-specific multiword expressions as a way to address data sparsity.  

\section*{Acknowledgements}

The authors would like to thank the Robert Bosch GmbH for providing access to the Vehicle Component Dataset and the DOMAIN corpus. We are grateful to Frank Hartmann, Andreas Körber, Patrick Reiser, and Boris Haselbach for valuable discussions and their support with the Bosch Vehicle Component Dataset.

We also thank Prisca Piccirilli, Neele Falk, and the SemRel group for constructive suggestions and feedback on versions of this work. Helena Schlipf was funded by the Robert Bosch GmbH. Annerose Eichel received funding by the Robert Bosch GmbH and the Hanns Seidel Foundation's Talent Program. 
\section*{Limitations}

When learning plausible materials for components in the vehicle repair domain, we build a varied set of query templates to probe PLMs and seek to aggregate obtained predictions in an optimal way. We are, however, fully aware that a set of queries that is optimal for machines is not necessarily the set that also makes perfect sense to humans.

As far as the transfer of the suggested approach to languages other than English is concerned, we call attention to the potential need to adapt the query templates, e.g., when working with languages that allow for more flexibility in word order such as Polish, Turkish, German, Hindi, or Finnish. Further, while it might be difficult to find well-curated domain-specific corpora in some languages, we show that using a customized version of Wikipedia of moderate size (approx. 225K sentences) in a given language represents a very viable alternative. Additionally, researchers could use a multilingual model or an adapter-based approach to navigate in-/output in other or multiple languages.

Our work focuses on RoBERTa as a backbone model, which has been shown to perform well in cloze task style settings. We also carried out initial experiments with newer models such as ELECTRA, not yielding desired results. We also conducted experiments with generative models such as GPT3. However, the output contained a lot of noise that did not aid our goals.
Moreover, the focus in this paper is devoted to the development of a novel contribution regarding the prompting and aggregation techniques as well as analyzing results rather than benchmarking a wide variety of models using an already existing method. While experiments with a wider variety of models represent an interesting future task, high-quality evaluation might be a bottleneck due to the availability and cost factor of domain experts. 

Regarding our model predictions, we have not yet attempted to detect and organize potential semantic relations between predictions, such as hypernym/hyponym relations between \textit{metal} and \textit{aluminium}. We leave organizing and utilizing such relations as potentially relevant milestone for connecting and tracing symptoms back to a cause for future work.
Furthermore, material predictions are not yet categorized along dimensions such as main and auxiliary materials or surface and inner materials, which might be of interest in the vehicle repair and material science domains. 

Finally, we would like to point out that all tested models and algorithms struggle with predicting plausible materials for intangible items which might, however, be more prevalent in other domains. Even if a model is able to predict correct material predictions, rating the plausibility of predictions that are not tangible but rather abstract materials such as \textit{data}, \textit{parameters} and \textit{functions} cannot be considered a trivial task for expert annotators and needs corresponding guidance.

\section*{Ethics Statement} \label{sec:ethics}
Two of the resources used in this work, the Vehicle Component Dataset and one of the two domain-specific corpora (DOMAIN) have been kindly provided to us. To provide insight into the nature of the data and foster reproducibility using comparable data, we make a sample of the Vehicle Component Dataset available (cf.~§\ref{sec:evaluation}). We also investigated comparable publicly available alternatives for DOMAIN and showed that leveraging a portion of the English Wikipedia customized to the domain of interest (WIKI) represents a viable substitute in case no custom corpus is available. Note that Wikipedia text content including Wikipedia dumps is licensed under both the Creative Commons Attribution-ShareAlike 3.0 License and the GNU Free Documentation License.

In the context of our evaluation task, we collected plausibility ratings from human participants. For this, the participants were provided an Informed Consent Letter with the name and the contact of the principal investigators; the title, purpose and procedure of the study; risks, benefits and compensation for participating in the study; confirmation of confidential anonymous data handling; and confirmation that participation in the study is voluntary. The Informed Consent Letter was signed by both the participant and the investigators before the participants took part in the study.

We use and adapt PLMs as provided and licensed under the Apache License 2.0 by \texttt{huggingface} \cite{wolf-etal-2020-transformers}. We acknowledge that material predictions learnt using the outlined approach are a product of unsupervised learning methods which might be prone to error. We point out that
predictions should be approved by an expert or flagged otherwise in case they are used in a downstream application to avoid potential risks including harm of objects or safety risks in case of incorrect repair procedures.

\bibliography{custom}
\bibliographystyle{acl_natbib}
\clearpage

\appendix
\section{Data}
\label{sec:data_distribution}

\begin{figure}[htpb!]
\centering
  \includegraphics[width=0.94\linewidth]{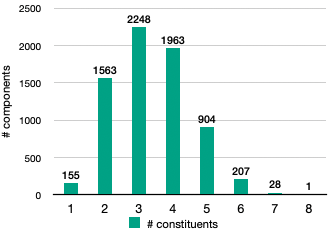}
  \caption{Distribution of multiword components in the Vehicle Component Dataset (in absolute numbers).}
  \label{fig:vcd_mwe_distribution}
\end{figure}

\paragraph{Dataset Distribution} Fig.~\ref{fig:vcd_mwe_distribution} shows the distribution of multiword component names in the Vehicle Component Dataset with the absolute number of vehicle components per number of constituent, e.g., the dataset comprises 1,563 multiword components with 2 constituents.

\section{Cloze Query Prediction and Processing}
\label{sec:cloze}

\subsection{Cloze Query Templates}
\label{subsec:cloze}
\paragraph{Verb Relation Variables} consist of, comprise, contain, be formed by, be formed of, be made up of, be made up from, be made of, be made out of, be composed of, be manufactured from, encompass, entail, include, involve, incorporate, be build of, be build with 

\paragraph{Cloze Query Statements} Table~\ref{tab:cloze_statements} presents an overview of the defined cloze query templates and provides an example of filled-in prompts for the vehicle component \textit{lamp}.

\subsection{Post-Processing}
\label{subsec:postprocessing}
\paragraph{PLM-based Predictions} Having created a given set of cloze queries, we feed the queries into a PLM and obtain the top-5 predictions. Based on our goals, we apply the following post-processing steps. We lower-case the predictions and filter out numbers, punctuation, 1-character tokens, as well as a list of standard stopwords obtained from \texttt{spacy} with nine customized stopwords including non-informative tokens such as \textit{material} and \textit{component}. Additionally, spelling variants such as \textit{aluminum} and \textit{aluminium} are merged. We make sure that a predicted token does not equal the singular or plural form of the full input component. 

\label{subsec:postprocessing}
\paragraph{Basilisk-based Predictions} We obtain $n$ predictions for each vehicle component using DOMAIN or WIKI for the Basilisk algorithm, 
and apply post-processing as described for the PLM predictions. 

\subsection{PLM and Basilisk Experiments: Computing Infrastructure and Model Information}
\label{subsec:domain_adaptation}
As Vanilla RoBERTa model, we use the \texttt{roberta-base} implementation from \texttt{huggingface} \cite{wolf-etal-2020-transformers} that comes with 125M parameters. We adapt this model to the vehicle repair domain. As DistilRoBERTa implementation, we use the \texttt{distilroberta-base} implementation from \texttt{huggingface} \cite{wolf-etal-2020-transformers} which has 82M parameters. See the \href{https://huggingface.co/transformers/v2.2.0/pretrained_models.html}{official huggingface documentation} for further details.
For all final experiments, including obtaining predictions from the various models, we use a single NVIDIA RTX A600 GPU. We adapted our models once (DOMAIN RoBERTa, WIKI RoBERTa, and DOMAIN DistilRoBERTa) and used the best model according to validation set performance, i.e. our domain-adapted models are the product of a single fine-tuning run. 

As a Basilisk implementation, we use the Java source code as provided by \citet{thelen-riloff-2002-bootstrapping}. We use \texttt{spaCy} \cite{spacy2020} for text preprocessing and dependency parsing.

\section{Evaluation}
\label{sec:component_sampling}
\subsection{Sampling Vehicle Components}
To test whether the head of a MWE preserves the meaning of the full MWE in our domain-specific data, we compare extracted material predictions using only the head of the MWE vs. all constituents (full). To evaluate results, we apply the following procedure when sampling vehicle components consisting of 2+ words. Full components are sampled as is along with the extracted results. Whenever only the head of a component is used, the results for only the head component are obtained. For the evaluation task, the head component is mapped back to the
original full component from the vehicle information dataset. For example, during sampling the head component \textit{sensor} gets mapped to the full 2-constituent component \textit{pressure sensor}. In this way, we can evaluate whether results extracted based on only the head \textit{sensor} are rated as plausible also for the more domain-specific full multiword expression \textit{pressure sensor}.

For the Basilisk-based approach, no results are extracted for one item of the 2-constituent full components and five items of the 3-constituent full components. These results were thus expanded with head component extraction results.
In one case, stopword removal
leads to an empty prediction list (\textit{alternator start charge current reduction, []}, using DOMAIN). The prediction is back-filled with a randomly sampled token from WIKI-based results for the head component (\textit{reduction}, [\textit{lithium}], using WIKI).

\subsection{Expert Evaluation} \label{app:expert_eval}

\paragraph{Evaluation Task Setup} The evaluation task was carried out in a remote setting using Google Forms. Annotators were provided detailed written guidelines including example questions and borderline decisions.
In case of questions, annotators had the option to contact the authors of the paper. 
The evaluation could be completed flexibly in the course of a week. Annotators could take as much as time as they needed for completing the evaluation (average time effort: \textasciitilde1.5 hours). Our three recruited annotators are based in Germany, male, and speak German as their first language. All annotators have 10+ years knowledge and continued education and training in English in general, as well as profound work experience with vehicle repair domain data in German and English.
Annotators received a representation allowance 
for their effort. Each annotator submitted one unique set of answers. 

The collected data does not include any information that names or uniquely identifies individual people or offensive content. We check for this by (i) not giving out data to annotators containing such information, and (ii) not asking for any of this information when collecting ratings regarding plausible material candidates for vehicle components, and (iii) manually inspecting collected data to confirm anonymity of annotators and potential other entities. Letters of Consent (cf. §\ref{sec:ethics}) are signed before participation and stored separately from the collected ratings.

\paragraph{Evaluation Results Analysis}
We evaluate a total of 2,342 material predictions from five different models and algorithms on a set of 100 vehicle components from the vehicle repair domain. For 99 components, we collect an average of 5.6 plausible material predictions valid for one or more models. For exactly one component, all annotators agree on \textit{I do not know the answer}.

Pair-wise inter-annotator agreement (IAA) on the level of vehicle components is shown in Table~\ref{tab:true_comps}. We calculate the percentage of components where two given annotators agree on \textit{at least one} material being plausible (i.e. accuracy on inter-annotator level) or where two given annotators agree that at least one material prediction is \textbf{not} plausible. This includes instances where up to four predictions are not plausible as well as instances where all five predicted material options are not rated as plausible (\textit{none of these}). We exclude instances where one or both annotators agree on the answer option \textit{I do not know the answer}.

\begin{table}[!htpb]
\centering
\small
\begin{tabular}{lr|rrr}
\toprule
Model             &
IAA  & A1::A2 & A2::A3 & A3::A1 \\ \midrule
Vanilla RB      & 1.00 & 1.00     & 1.00     & 1.00      \\
DOMAIN RB       & 1.00 & 0.99     & 1.00     & 1.00      \\
WIKI RB         & 1.00 & 1.00     & 1.00     & 1.00      \\
Basilisk        & 0.93 & 0.97     & 0.91     & 0.91      \\
DOM. DistilRB & 1.00 & 1.00     & 1.00     & 1.00      \\ \bottomrule
\end{tabular}
\caption{Proportion of components where two A$n$ agree on at least one plausible material or where A$n$s agree on at least one material prediction being \textit{not} plausible.}
\vspace{-3mm}
\label{tab:true_comps}
\end{table}



\section{Results and Discussion}
\label{sec:results_discussion}

\subsection{Statistical Analysis: Number of Cloze Queries and Component Accuracy} \label{subsec:statistical_analysis}

We statistically analyze the relationship between the number of cloze queries triggering predictions for a given component and the actual accuracy values on component level.\footnote{As no cloze queries are generated for the Basilisk algorithm, we only investigate PLM results.} For this, we first perform a correlation analysis to see whether the number of activated queries is correlated to component accuracy, i.e., whether 1+ of the predictions for this component is rated plausible by 1+ expert annotator. Results are reported in Table~\ref{tab:statistical_analysis} with the investigated models on the left and Pearson's correlation coefficients in the first column. The obtained results indicate that the relationship between the number of activated queries and observed component accuracy is strongly correlated (coefficient $>$0.5) for the Vanilla and WIKI RB models, and moderately correlated (coefficient $>$0.3) for DOMAIN RB and DOMAIN DistilRB.

\begin{table}[!htpb]
\centering
\small
\begin{tabular}{l|r|r}
\toprule
\multicolumn{1}{l}{Model}            & \multicolumn{1}{l}{Pearson's Coefficients} & \multicolumn{1}{l}{$R^{2}$} \\ \midrule
Vanilla RB  & 0.51$^{***}$  & 0.26$^{***}$                                         \\
DOMAIN RB   & 0.49$^{***}$   & 0.24$^{***}$                                                \\
WIKI RB     & 0.58$^{***}$ & 0.34$^{***}$                                                   \\
DOM. DistilRB & 0.40$^{***}$  & 0.16$^{***}$                                                     \\ \bottomrule
\end{tabular}
\caption{Pearson's correlation coefficients for the number of activated queries and component accuracy. All coefficients are significant with a p-value $<0.01$ (***).}
\label{tab:statistical_analysis}
\vspace{-3mm}
\normalsize
\end{table}

To see whether these correlations are in fact directed, we employ linear regression modeling with the number of queries as predictor and component accuracy as outcome variable. Results are presented in Table~\ref{tab:statistical_analysis} with investigated models on the left and $R^{2}$ values in the second column. The relatively low $R^{2}$ values suggest that the relationship between the number of activated queries and observed \textit{component} ($\neq$\textit{material}) accuracy might more strongly rely on other factors than the number of queries activated for a given component.   

\subsection{$n$-Constituent Component Analysis} \label{subsec:n_tokens}

\begin{figure}[htpb!]
\centering
  \includegraphics[width=\linewidth]{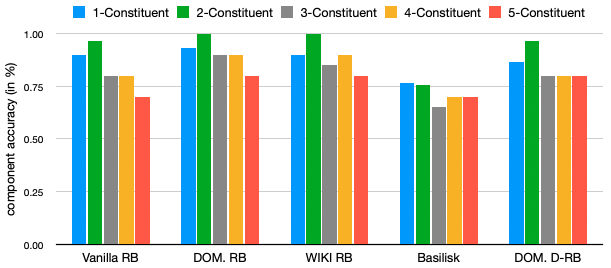}
  \caption{Results regarding $n$-constituent performance on component accuracy level (in \%).}
  \label{fig:ntokens_analysis}
\vspace{-3mm}
\end{figure}

Results regarding MWEs with $n$ constituents are shown in Fig.~\ref{fig:ntokens_analysis}.

Model performance is presented on component accuracy level. It can be observed that 2-constituent components receive the highest number of plausible material annotations, followed by simplex components, consisting of only one constituent. 4-constituent components outperform the shorter 3-constituent and the longest investigated components, namely 5-constituent components. The results indicate that (domain-)specificity beats brevity when learning plausible materials for vehicle components. However, our findings also suggest that there might be a threshold separating a beneficial number of constituents (2) from a number pointing towards a detrimental degree of specificity (3+). 

\subsection{Prediction Pool Size Analysis} \label{subsec:pool_size_analysis}

We present results regarding the size of the pool from which predictions are drawn for aggregation in Table~\ref{tab:pool_size_analysis}. 
\begin{table}[!h]
\centering
\small
\begin{tabular}{lrr}
\toprule
            & Top-10 Pool   & Top-5 Pool    \\ \midrule
Hits@1      & 0.54          & 0.37          \\
Hits@2      & 0.45          & 0.56          \\
Hits@3      & 0.45          & 0.42          \\
Hits@4      & 0.50          & 0.42          \\
Hits@5      & 0.41          & 0.31          \\ \midrule
Comp. Acc@5 & 0.85          & \textbf{0.86} \\ \midrule
Total Hits@5     & \textbf{0.47} & 0.42          \\ \bottomrule
\end{tabular}
\caption{Correct predictions (Hits@k) per top-k predictions (x/100), component accuracy (x/100), and total correct predictions per top-5 predictions (Total Hits@5).}
\label{tab:pool_size_analysis}
\vspace{-3mm}
\normalsize
\end{table}
The percentage of correct predictions per top-k predictions are depicted as Hits@1-5. Here, using a larger pool leads to a gain of 18\% more plausible materials as the top-1 prediction. While accuracy on component level stays basically the same (top-10 pool: 0.85\%, top-5 pool: 0.86\%), an increase of 5\% can be observed when building the final top-5 predictions from a larger pool of predictions.

\end{document}